# Boundary Evolution Algorithm for SAT-NP

Zhaoyang Ai[①], Chaodong Fan[②], Yingjie Zhang[③], Huigui Rong[④], Ze'an Tian[③], Haibing Fu[③] [1]

**Abstract:**

A boundary evolution Algorithm (BEA) is proposed by simultaneously taking into account the bottom and the high-level crossover and mutation, ie., the boundary of the hierarchical genetic algorithm. Operators and optimal individuals based on optional annealing are designed. Based on the numerous versions of genetic algorithm, the boundary evolution approach with crossover and mutation has been tested on the SAT problem and compared with two competing methods: a traditional genetic algorithm and another traditional hierarchical genetic algorithm, and among some others. The results of the comparative experiments in solving SAT problem have proved that the new hierarchical genetic algorithm based on simulated annealing and optimal individuals (BEA) can improve the success rate and convergence speed considerably for SAT problem due to its avoidance of both divergence and loss of optimal individuals, and by coronary, conducive to NP problem. Though more extensive comparisons are to be made on more algorithms, the consideration of the boundary elasticity of hierarchical genetic algorithm is an implication of evolutionary algorithm.

[1] [②]Institute of Language and Cognition, College of Foreign Languages, Hunan University, China
[②]Institute of Information and Engineering, Xiangtan University, China , Corresponding author.
[②]Institute of Information and Engineering, Hunan University, China
[③]College of Physics and Electronics, Hunan University, China

## 1 Boundary effect of evolution

Evolution is reflected in the boundary effect of the species community. Evolution begins on the boundary between two neighboring species. On the boundary of the two neighboring species communities, there are clashes and changes because of crossover. The change of a few individuals is called mutation, while the change of the whole population can be an evolution, because there might be a distinguished boundary between them. And then a new species community evolves and forms. For example, monkey group A has one or two individual monkeys who can make fire. This change is only a mutational change at most. Monkey group B has no individual capable of making fire. Then forms another boundary which can divide two different species groups. If all the population of monkey group A are capable of making fire, then a new species forms. This shows that the local optimal survivability eventually becomes a global optimal solution for its probabilistic jumps of mutations. This change is an evolutionary change and monkey group A has evolved. In a word, a successful evolution is a spread of the mutation into the whole population. The mutation of several individuals among a community is only a mutation, instead of an evolution. An evolution is realized only when a whole community is evolutionized, beginning from the boundary mutations, which is an effect from the two extremes, ie., both the bottom and the top.

This is consistent with the principle of annealing for the hierarchical evolution. The annealing principle requires that the annealing effects holistically. The boundary for the temperature is the boundary for the whole population. The range of this kind of boundary change is called the boundary elasticity. The annealing is not individual but integrated and holistic. Therefore, the evolutionary algorithm to use the annealing method is to implement the boundary elasticity with mutation operators. This optimization is a very good idea for evolutionary algorithms. The local optimal solutions tend to be eventually global optimal solutions for its probabilistic jumps. Because of the whole population's mutational change, the boundary mutation evolves into optimal operator. Its efficiency is to improve the convergence, and the convergence of boundary evolves into a whole new population. All this is the result of the boundary effect of the evolution. The consideration of the boundary elasticity of hierarchical genetic algorithm is an implication of evolutionary algorithm. The BEA is an option for the evolution algorithm.

## 2 Basics

The satisfiability problem (SAT) of Conjunction Normal Form (CNF) is a famous NP problem. The SAT problem has been proved to be among the NP problems (Gu Wen-xiang, 2012). Many NP problems can be transformed into SAT problems, which is called SAT-NP. Therefore, the research on SAT problem is of great theoretical value. Some practical problems such as protein folding, computer vision, computer networking and scheduling can also be changed into SAT problems (Fox G. 2001). In the field of Electronic Design Automation（EDA）of Integrated Circuits, SAT can be used effectively in Automatic Test Pattern Generation (ATPG), time-sequence analysis, logic checking and equivalence checking, and among others (Jing Ming-e et al. 2008). As a result, the SAT research is of great practical value as well.

Until now there have been two mainstream SAT solvers: Complete Algorithm and Incomplete Algorithm (Lin Fen et al., 2009). The complete algorithm (Guo Y, Zhang B, Zhang C, 2012) has developed from DPLL (Davis M, Putnam H. 1960) ( Davis M. et al.,1960). It can theoretically prove the unsatisfiability or obtain solutions to SAT problem, but cannot obtain the solutions when run within an appropriate period of time. The incomplete algorithm, however, cannot necessarily find optimal solutions, but it solves problems fast, more efficient than the complete algorithm in most cases.

Therefore, the research on SAT problem has mainly turned its direction towards efficient and practical incomplete algorithm. The incomplete algorithm consists of two parts. One is the algorithm adopting local search. (Selman B et al,1994) ( Brys T et al.,2013) (Lü Z, et al.,2012) (Bouhmala N and Salih S., 2012). Selman B (1994) solves the satisfiability problem and the maximum satsifiability by using the improved local search under the noise strategy. Brys T. (2013) solves the satisfiability problem of fuzzy logic by using the local search and restarting strategy. Lü Z (2012) solves the maximum satisfiability by using the adaptive local search based on memory. Bouhmala N(2012) solves the maximum satisfiability problem using multilevel Tabu search. The other is the one adopting evolution (Folino G. et al.,2001) (Sun Qiang et al. 2008) (Li Yang-yang and Jiao Li-cheng,2007) (**Zhao C**.et al., 2012) (**Zhao C. et al., 2011).** Zhao C (**Zhao C**.et al., 2011,2012) adopts algorithm based on BP. Folino (2001) adopts CGWASAT algorithm. Sun Qiang (2008) adopts the annealing genetic algorithm to solve SAT problem. Li Yang-yang (2007) uses quantum immune cloning algorithm to solve SAT problem.

Generally speaking, the incomplete algorithms mentioned above are either limited to local optimum or subject to low convergence speed. This paper, however, assumes that there is a boundary effect in hierarchical genetic algorithm, and if we tend to the boundary of the hierarchical algorithm, ie., the bottom and the high-level, the algorithm can be improved considerably. In this paper crossover and mutation operators are designed based on simulated annealing in the frame of hierarchical genetic algorithm. A new high-level selection operator has been designed to prevent loss of optimal individuals, and then the improved hierarchical genetic algorithm is applied to SAT solution. The experimental results show that the improved hierarchical genetic algorithm has prominently better performance in success rate, iterations and among others.

### 2.1 Description of SAT problem

There are different descriptions for SAT problem. Fan Chao-dong (2011) provides the following definition:

**Definition 1** For a given propositional variable set,
$$V=\{x_1,x_2,\ldots,x_n\} \qquad (1)$$
$x_i \in \{0,1\}$，$1 \leq i \leq n$，0 stands for *false*，and 1 stands for *true*. The value of any of the propositional variables $x_1,x_2,\ldots,x_n$ is called a truth value assignment.

**Definition 2** The word stands for *any variable $x_i$* in

$V$, or a non-variable $\neg x_i$. The clause stands for the disjunction of some words, and the conjunction normal form (CNF) stands for the conjunction of certain clauses. SAT problem is to decide whether there is a truth assignment for $\{x_1,x_2,\cdots,x_n\}$, which makes CNF true.

## 2.2 Simulated annealing

Simulated annealing (SA) was first proposed by Metropolis et al in 1953. It was based on the similarity in physics between the annealing process of solid matters and the general combinatorial optimization. According to Wang Ling et al. (2001) the simulated annealing algorithm searches randomly in the solution space, characteristic of sudden jumps of probability, for the global optimal solution to the target function, while the temperature parameters are declining from an initial temperature. The local optimal solutions eventually tend to be global optimal solutions for its probabilistic jumps. As a general optimal algorithm, simulated annealing has been widely used in different fields and promises a good future. The steps of the algorithm can be found as follows (Yi L et al, 2009):

Step1: The initial temperature is $t=t_0$, and the randomly generated initial state is $S=S_0$, and let $k=0$;

Step2: The random stirring generates a novel state $S_1$. Run $\triangle S = S_1 - S_0$;

Step3: If $\min\{1,\exp(-\triangle S/t_k)\} \geq \text{random}[0,1]$, then $S_1$ is accepted as the present state;

Step4: Annealing $t_{k+1}=\text{update}(t_k)$, let $k=k+1$;

Step5: If there is no results, then return to Step2. Otherwise the algorithm is ended and gives results.

**2.3 Hierarchical Genetic Algorithm**

The hierarchical genetic algorithm is an improvement on traditional genetic algorithm. It is based on the idea that the sub-populations independently conduct their genetic operations for some time and then can obtain some optimal models at some particular positions of the individual chains. Then novel individuals of various optimal models will be obtained by means of the high level genetic algorithm. Its algorithm has the following steps (Liu Hai-di, 2008):

Step1: Initialize $N$ sub-populations;

Step2: Let the $N$ sub-populations conduct separately certain algebraic genetic algorithms;

Step3: Decide whether the algorithm satisfies the conditions for ending the calculation. If the conditions are satisfied, the calculation ends; otherwise, it continues;

Step4: Write separately the $N$ resulted populations and their mean adaptive value into $R[1\ldots N, 1\ldots n]$ and $A[i]$;

Step5: Run selective, crossover and mutation operations in $R[1\ldots N, 1\ldots n]$;

Step6: Return to Step2 to start over genetic operations.

## 3 Design of improved Operators

### 3.1 Crossover and Mutation operators based on annealing

The traditional hierarchical genetic algorithm often conducts classic algorithm at the bottom. In the classic genetic algorithm the crossover and mutation operators cross over and mutate randomly within certain probabilities. In this situation the evolution becomes slow, and some optimal individuals can be destroyed, causing degradation to some degrees in the later stages. To overcome the degradation and other problems of traditional genetic algorithm, this paper proposes crossover and mutation operators based on simulated annealing.

**Definition 3** Crossover operators based on simulated annealing

Suppose $A,B$ are selected individuals for crossover, $f(x)$ stands for the adaptive value of $x$, and $T$ stands for the annealing temperature.

$$t_1=\max\{f(A),f(B)\} \quad (2)$$

Suppose $C,D$ are the individuals generated by the crossover of $A,B$

$$t_2=\max\{f(C),f(D)\} \quad (3)$$

The annealing-based crossover operator proposes the criteria for the crossover:

(1) If $t_2 \geq t_1$, then the crossover is successful.

(2) If $t_2 < t_1$, then a random number $z$ within $0\sim 1$ is generated, and if

$$\exp(-(t_2-t_1)/T) > z \quad (4)$$

then the crossover is also successful; otherwise the crossover fails.

**Definition 4 Mutation operators based on simulated annealing**

Suppose A is the individual selected for mutation, $f(x)$ represents the adaptive value of $x$, and $T$ stands for the annealing temperature; Suppose $A'$ is the mutated individual of $A$, the simulated-annealing-based mutation operator proposes the criteria for mutation operation as follows:

(1) If $f(A')>f(A)$, then the mutation succeeds.
(2) If $f(A')<f(A)$, then a random number $z$ within 0~1 is generated, and if

$$\exp(-(f(A)-f(A))/T)>z \qquad (5)$$

the mutation is also successful ; otherwise the mutation fails.

Therefore, the post-annealed genetic operation is to be accepted if the solution generated by the improved genetic operator based on simulated annealing is better than the previous solution; the previous genetic operation is to be accepted within certain probability if the solution generated by the improved operator is poorer than the previous solution. To this effect the simulated annealing operators can improve the effeciency of the algorithm by avoiding the deffeciency of random operation by traditional algorithm.

## 3.2 The high-level selection operator based on optimal individuals

The high-level operators of traditional hierarchical genetic algorithm conduct selection operation only based on the mean adaptive value of separate bottom populations. This way the bottom population is likely to be excluded by the traditional hierarchical genetic algorithm because the bottom population's mean adaptive value is very low when the population that contains the optimal individual also contains a lot of individuals with low adaptive value. Evidently this is not beneficial for the optimization of algorithm. Consequently a novel high-level selection operator is designed based on optimal individuals.

**Definition 5 High-level Selection Operator based on Optimal individuals**

Suppose $g_i$ represents the adaptive value of the $i$th population, and $r_i$ stands for the mean adaptive value of the $i$th population, then the adaptive value of the high level genetic operator is $\alpha g_i + \beta r_i$, and the probability of the high level selection is:

$$p_i = \frac{\alpha g_i + \beta r_i}{\sum_{i=1}^{n}(\alpha g_i + \beta r_i)} \qquad (6)$$

The value range of $\alpha$、$\beta$ is 0~1. Consequently as is found, the high level selection operators based on optimal individuals of the bottom population can offset the deficiency of traditional hierarchical genetic algorithm.

## 4 The boundary evolution approach for hierarchical genetic algorithm

### 4.1 Steps for the algorithm

Under the frame of traditional hierarchical genetic algorithm, our boundary evolution approach for hierarchical algorithm adopts annealed crossover and mutation operators at the bottom genetic algorithm, and adopts the high level operators based on optimal individuals at the high level genetic algorithm. Here are the steps to follow:

(1) Initialize the parameters of the algorithm;
(2) Randomly generate $N$ sub-populations of scale $n$;
(3) The $N$ sub-populations run bottom genetic algorithm for times independently based on the above-mentioned annealed crossover and mutation operators;
(4) The high level genetic algorithm runs selection operation based on the above-mentioned selection operator and optimal individuals to obtain the next generations of population;
(5) Crossover and mutate at the high level genetic algorithm;
(6) Decide whether conditions for termination are satisfied. (The conditions for termination include that the optimal individuals satisfy all clauses, or that the algorithm reaches the maximum iterations.) If the conditions are satisfied, then give the results, and the algorithm ends; otherwise, return to step (3).

### 4.2 The Algorithm Flow

The flow of the boundary evolution algorithm is demonstrated in the graph as found in Figure 1.

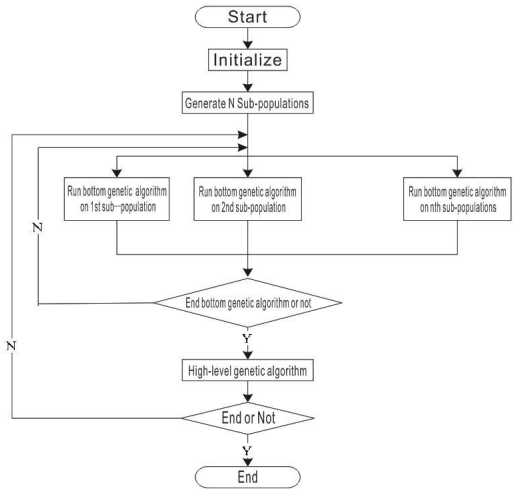

Figure 1 Algorithm flow

## 5 Experimental test and Result analyses

### 5.1 Tests on SAT problems and their result analyses

To testify the effectiveness of the proposed Improved Annealed hierarchical genetic algorithm (BEA), we use it to solve SAT problems. Out of convenience for analyses, it is compared with the traditional hierarchical genetic algorithm that adopts elitist crossovers (HGA), and the traditional hierarchical genetic algorithm that adopts both elitist crossovers and selection operators based on optimal individuals (BIHGA). The test environment is: PC with Pentium(R) 4 3.00GHZ CPU，512MB memory. The operation system: Windows XP.

Table 1 Parameters for algorithm

| Parameters | Setting values |
|---|---|
| Sub-population amounts | 4 |
| Sub-population scale | 5 |
| Algebra for Sub-population evolution | 50 |
| Algebra for High level maximum evolution | 10000 |
| Initial temperature | Number of clauses *length of clauses |
| Cooling factor | 0.95 |
| Bottom and high level mutation rate | 0.0001 |
| $\alpha$ | 0.5 |
| $\beta$ | 0.5 |

When the ratio between the clause and the variable $k = m/n$ approaches 4.3, there is neither too much nor too little constraint for SAT problems. The probability for satisfiability is equal to that for unsatisfiability. This kind of problem is difficult to solve (Hu Xian-wei, 2012). This paper adopts Random Model 3-SAT with $k \approx 4.3$ as test sample.

In the experiment the parameters for the algorithm are shown in table 1.

Each group of test data is used 10 times. The mean algebra refers to the algebra for the 10 times' mean high-level evolution, and the average success rate refers to the ratio between the times of success and the total times of problem solving. The experiments' data are shown in table 2.

The results of the experiments show that BIHGA, in comparison with HGA, has prevented loss of optimal individuals to some degree for its high-level operators and optimal individuals. In addition, the convergence speed has increased due to the preservation of the optimal individuals. What's more, BEA has avoided randomness in crossover and mutation and greatly increased convergence speed and success rate of problem-solving because of its crossover and mutation operators based on simulated annealing.

Table 2 Comparative Performance of BEA, HGA. BIHGA

| CNF | | Mean algebra for Evolution | | | Average Success Rate | | |
|---|---|---|---|---|---|---|---|
| m | n | BEA | BIHGA | HGA | BEA | BIHGA | HGA |
| 91 | 20 | 3.8 | 4.6 | 5.5 | 100% | 100% | 100% |
| 215 | 50 | 4.9 | 1041.7 | 2855 | 100% | 100% | 100% |
| 325 | 75 | 5.0 | 8641 | 9514 | 100% | 20% | 10% |
| 430 | 100 | 5.7 | - | - | 100% | 0 | 0 |

To further testify the effectiveness of BEA, comparison has been made between the proposed Improved Annealed Hierarchical Genetic Algorithm (IAHIGA=BEA) based on annealed crossover and optimal-individual at high level selection and the traditional hierarchical genetic algorithm(THGA). The test results are shown in Figure 2 and Figure 3 separately. (The horizontal axis represents the

algebra for evolution, and the vertical axis stands for the adaptive value of the optimal individuals.)

Figure 2 Comparison of the three algorithms in solving Problem 1(IAHGA=BEA)

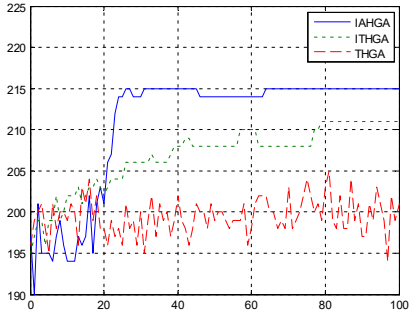

Figure 3 Comparison of the three algorithms in solving Problem 2(IAHGA=BEA)

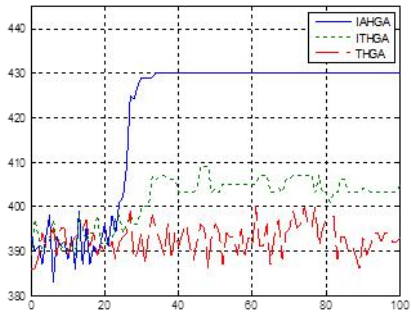

Figure 2 shows the comparative results of solutions to problem 1 (The SAT problem with 50 variables and 215 clauses). Figure 3 shows the comparative results of solution to problem 2 (This is an SAT problem with 100 variables and 430 clauses). Obviously, BEA converges faster than THGA due to the annealed crossover operators and high-level selection operators based on optimal individuals. BEA has a far greater solution rate due to the annealed crossover and mutation operators.

## 5.2 Comparative Experiments on other optimal algorithms

In order to further verify the BEA., this section selects 164 satisfiables of the 24 sub-categories from SATLIB repository for test. The test results were compared with the literature **Guo Ying** (Guo Ying et. al 2014) GASAT and ACOSAT. The parameter setting follows **Guo Ying** (Guo Ying et. al 2014). The experimental data of GASAT and ACOSAT were selected from **Guo Ying** (Guo Ying et. al 2014). The results from the comparison are shown in table 3. From table 3, it is found that GASAT is of the poorest performance among the three, except that it achieves a relatively good results for problem $T_5$. BEA can solve all the SAT problems except $T_{16}$. In comparison with GASAT and ACOSAT, BEA enjoys obvious advantage in average optimal value and average success rate. The results of the comparison have further verified the effectiveness of BEA, which can be used to solve the problem of SAT-NP, and it has great prospects for wide applications.

Table 3 Comparative Performance of BEA, GASAT, ACOSAT

| Problem | Average Optimal Value | | | Average Success Rate | | |
|---|---|---|---|---|---|---|
| | GSAT | ACOSAT | BEA | GSAT | ACOSAT | BEA |
| $T_1$ | 1.3 | 0 | 0 | 0 | 100% | 100% |
| $T_2$ | 1.1 | 0 | 0 | 0 | 100% | 100% |
| $T_3$ | 3.4 | 0 | 0 | 0 | 100% | 100% |
| $T_4$ | 2.7 | 0 | 0 | 0 | 100% | 100% |
| $T_5$ | 2.0 | 0 | 0 | 100% | 100% | 100% |
| $T_6$ | 1.6 | 0 | 0 | 0 | 100% | 100% |
| $T_7$ | 5.8 | 0 | 0 | 0 | 100% | 100% |
| $T_8$ | 2.6 | 0 | 0 | 0 | 100% | 100% |
| $T_9$ | 5.4 | 0 | 0 | 0 | 100% | 100% |
| $T_{10}$ | 8.5 | 0 | 0 | 0 | 100% | 100% |
| $T_{11}$ | 1.6 | 0 | 0 | 40% | 100% | 100% |
| $T_{12}$ | 1.0 | 0 | 0 | 30% | 100% | 100% |
| $T_{13}$ | 2.4 | 0 | 0 | 20% | 100% | 100% |
| $T_{14}$ | 3.4 | 0 | 0 | 0 | 100% | 100% |
| $T_{15}$ | 5.1 | 0 | 0 | 0 | 100% | 100% |
| $T_{16}$ | 4.4 | 3.2 | 1.7 | 0 | 95% | 99% |
| $T_{17}$ | 7.1 | 5.7 | 0 | 0 | 85% | 100% |
| $T_{18}$ | 6.5 | 6.8 | 0 | 0 | 90% | 100% |
| $T_{19}$ | 7.8 | 7.2 | 0 | 0 | 85% | 100% |
| $T_{20}$ | 7.7 | 6.5 | 0 | 0 | 70% | 100% |

The results of the comparative experiments in solving SAT problem have proved that the new hierarchical genetic algorithm based on simulated annealing and optimal individuals can improve the success rate and convergence speed considerably for SAT problem due to its avoidance of both divergence and loss of optimal individuals, and by coronary, conducive to NP problem. Though more extensive comparisons are to be made on more algorithms, the consideration of the boundary elasticity of hierarchical genetic algorithm is an implication of evolutionary algorithm.

# 6 Conclusion

Traditional hierarchical genetic algorithm is defective for its divergence and slow convergence because it adopts classic crossover and mutation operators at the bottom algorithm. It is also easy to lose optimal solutions because the high level selection operators mostly adopt the mean adaptive value of the sub-population as criterion. However, the simulated annealing algorithm is a global optimal algorithm based on the sudden jumps of probability in the annealing process of solid matters. Therefore, it can increase the convergence speed and avoid the divergence of the bottom algorithm by improving the classic crossover and mutation operators based on simulated annealing. This is because it follows the boundary elasticity principle to consider the boundary mutation and the holistic evolution to realize the optimized community evolution. The improved high-level selection operators and optimal individuals can effectively prevent the optimal individuals from getting lost. The application of the boundary evolution algorithm and its experimental results of solving SAT problems have proved that the proposed algorithm has obviously improved the success rate of solution and the convergence speed, and has overcome the inherited deficiency of traditional hierarchical genetic algorithm in solving SAT-NP problems. The boundary evolution algorithm is to be further developed and tested for an important principle for evolution algorithm.

**Acknowledgements** We acknowledge the support of National Natural Science Foundation of China (61174140), Hunan Provincial Natural Science Foundation of China (13JJA002), Hunan Provincial Education Foundation (Y00012), and Doctoral Foundation of Education Ministry of China (20110161110035), Hunan Provincial Social Science Grant 2014(XJK015AGD003); Hunan Provincial Education Science Key Grant 2013(14YBA084).